\documentclass[letterpaper]{article} 
\usepackage{aaai23}  
\usepackage{times}  
\usepackage{helvet} 
\usepackage{courier}  
\usepackage[hyphens]{url}  
\usepackage{graphicx} 
\usepackage{natbib}  
\usepackage{caption} 
\usepackage[utf8]{inputenc} 
\usepackage[T1]{fontenc}    
\usepackage{hyperref}       
\usepackage{url}            
\usepackage{booktabs}       
\usepackage{amsfonts}       
\usepackage{nicefrac}       
\usepackage{microtype}      
\usepackage{xcolor}         
\usepackage{algorithm}
\usepackage{algorithmic}
\usepackage{amssymb}
\usepackage{amsmath}
\usepackage{amssymb}
\usepackage{mathtools}
\usepackage{amsthm}
\usepackage{wrapfig}
\usepackage{makecell}

\usepackage[capitalize,noabbrev]{cleveref}

\usepackage{pifont}
\newcommand{\cmark}{\color{green}{\ding{51}}} 
\newcommand{\xmark}{\color{red}{\ding{55}}}

\usepackage{newfloat}
\usepackage{listings}
\DeclareCaptionStyle{ruled}{labelfont=normalfont,labelsep=colon,strut=off} 
\floatstyle{ruled}
\newfloat{listing}{tb}{lst}{}
\floatname{listing}{Listing}

\usepackage[textsize=tiny]{todonotes}
\newcommand{\ts}{\textsuperscript}
\usepackage[linesnumbered,ruled,vlined,algo2e]{algorithm2e}
\usepackage{enumitem}
\usepackage{graphicx}
\usepackage{wrapfig}

\definecolor{Gray}{gray}{0.9}
\definecolor{LightCyan}{rgb}{0.88,1,1}
\renewcommand{\paragraph}[1]{\vspace{-0.5ex}\textbf{#1}}

\newcommand{\ie}{i.e.\xspace}
\newcommand{\eg}{e.g.\xspace}

\usepackage{amssymb}
\usepackage{amsmath,amsfonts}
\usepackage{amsopn}
\usepackage{bm} % bold symbol
\usepackage{multirow}
\usepackage{flushend}
\usepackage{tabularx}
\usepackage{tikz}

\newcommand{\FE}{\method{FE}\xspace}

\newcommand{\vct}[1]{\boldsymbol{#1}} % vector
\newcommand{\mat}[1]{\boldsymbol{#1}} % matrix
\newcommand{\field}[1]{\mathbb{#1}}
\newcommand{\R}{\field{R}} % real domain
\newcommand{\ProbOpr}[1]{\mathbb{#1}}
\newcommand{\expect}[2]{%
\ifthenelse{\equal{#2}{}}{\ProbOpr{E}_{#1}}
{\ifthenelse{\equal{#1}{}}{\ProbOpr{E}\left[#2\right]}{\ProbOpr{E}_{#1}\left[#2\right]}}} % Expectation: syntax: E{1}{2} = E_1[2], E{}{2}=E[2], E{1}{} = E_1
\newcommand{\vb}{\vct{b}}

\newcommand{\vz}{{\vct{z}}}
\newcommand{\vv}{\vct{v}}

\newcommand{\mA}{\mat{A}}

\newcommand{\mS}{\mat{S}}

\newcommand{\mI}{\mat{I}}

\newcommand{\sS}{\mathcal{S}}

\newcommand{\sL}{\mathcal{L}}

\newcommand{\sZ}{\mathcal{Z}}
\newcommand{\eat}[1]{}
\newcommand{\method}[1]{\textsc{#1}}

\SetKwInput{KwInput}{Input}                % Set the Input
\SetKwInput{KwOutput}{Output}              % set the Output
\SetKwInput{KwInit}{Initialize}              % set the Output

\title{Learning Fractals by Gradient Descent}
\author{
    Cheng-Hao Tu\equalcontrib,
    Hong-You Chen\equalcontrib,
    David Carlyn,
    Wei-Lun Chao
}
\affiliations{
Department of Computer Science and Engineering, The Ohio State University\\
\texttt{\{tu.343, chen.9301, carlyn.1, chao.209\}@osu.edu}}

\usepackage{bibentry}

\begin{document}

\maketitle

\begin{abstract}
Fractals are geometric shapes that can display complex and self-similar patterns found in nature (\eg, clouds and plants).
Recent works in visual recognition have leveraged this property to create random fractal images for model pre-training.
In this paper, we study the inverse problem --- \emph{given a target image (not necessarily a fractal), we aim to generate a fractal image that looks like it.}
We propose a novel approach that learns the parameters underlying a fractal image via gradient descent. We show that our approach can find fractal parameters of high visual quality and be compatible with different loss functions, opening up several potentials, \eg, learning fractals for downstream tasks, scientific understanding, etc. 
\end{abstract}

\section{Introduction}
\label{intro}

A fractal is an infinitely complex shape that is self-similar across different scales. It displays a pattern that repeats forever, and every part of a fractal looks very similar to the whole fractal. Fractals have been shown to capture geometric properties of elements found in nature~\cite{mandelbrot1982fractal}. In essence, our nature is full of fractals, \eg, trees, rivers, coastlines, mountains, clouds, seashells, hurricanes, etc. 
 
\begin{figure*}[t]
\centering
\includegraphics[width=0.9\textwidth]{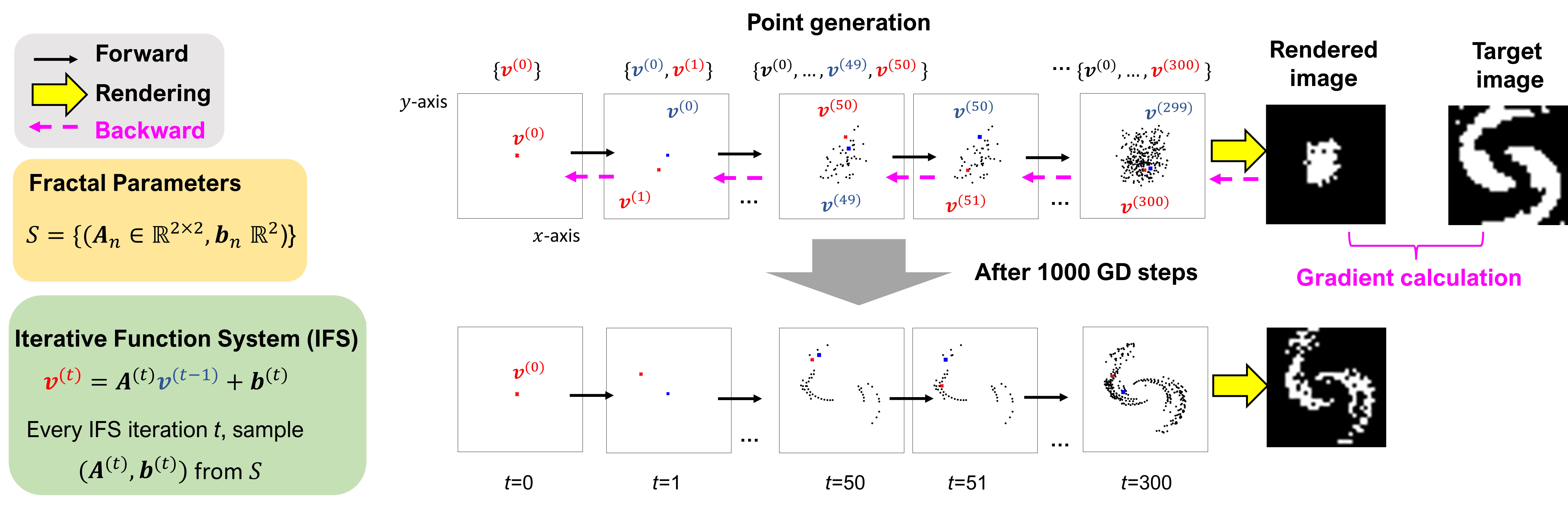}
\vskip -2pt
\caption{\small \textbf{Overview of our approach}, which aims to learn the fractal parameters $\sS$ so that the generated point sequence by IFS (\ie, $\{\vv^{(0)},\cdots,\vv^{(300)}\}$), after being rendered onto an image, is close to the target image. We learn $\sS$ by gradient descent (GD). As shown, after $1000$ GD steps, we can learn high-quality $\sS$ to generate images that are visually close to the target one.
}
\vskip -10pt
\label{fig:render}

\end{figure*}

Despite its complex shape, a fractal can be generated via well-defined mathematical systems like iterated function systems (IFS)~\cite{barnsley2014fractals}. 
An IFS generates a fractal image by ``drawing'' points iteratively on a canvas, in which the point transition is governed by a small set of $2\times 2$ affine transformations (each with a probability). Concretely, given the current point $\vv^{(t)}\in\R^2$, the IFS randomly samples one affine transformation with replacement from the set, and uses it to transform $\vv^{(t)}$ into the next point $\vv^{(t+1)}$. This stochastic step is repeated until a pre-defined number of iterations $T$ is reached. The collection of $\{\vv^{(0)}, \cdots, \vv^{(T)}\}$ can then be used to draw the fractal image (see \autoref{fig:render} for an illustration). \emph{The set of affine transformations thus can be seen as the parameters (or ID) of a fractal.}

Motivated by these nice properties, several recent works in visual recognition have investigated generating a large number of fractal images to facilitate model (pre-)training, in which the parameters of each fractal are randomly sampled~\cite{kataoka2020pre,Anderson_2022_WACV}. For instance, \citet{kataoka2020pre} sampled {multiple sets} of parameters and treated each set as a fractal class. Within each class, they leveraged the stochastic nature of the IFS to generate fractal images with intra-class variations. They then used these images to pre-train a neural network in a supervised way without natural images and achieved promising results.

In this paper, we study a complementary and inverse problem --- \emph{given a target image that is not necessarily a fractal, can we find the parameters such that the generated IFS fractal image looks like it?} We consider this problem important from at least two aspects. First, it has the potential to find fractal parameters suitable for downstream visual tasks, \eg, for different kinds of image domains. Second, it can help us understand the limitations of IFS fractals, \eg, what kinds of image patterns the IFS cannot generate.

We propose a novel gradient-descent-based approach for this problem (see \autoref{difffractal}). Given a target image and the (initial) fractal parameters, our approach compares the image generated by the IFS to the target image, back-propagates the gradients w.r.t.~the generated image \emph{through the IFS}, and uses the resulting gradients w.r.t.~the parameters to update the fractal parameters. The key insights are three-folded. First, we view an IFS as a special recurrent neural network (RNN) with identity activation functions and stochastic weights. The weights in each recurrent step (\ie, one affine transformation) are not the same, but are sampled with replacement from shared parameters. This analogy allows us to implement the forward (\ie, image generation) and backward (\ie, back-propagation) passes of IFS using deep learning frameworks like PyTorch~\cite{paszke2019pytorch} and TensorFlow~\cite{abadi2016tensorflow}. Second, the IFS generates a sequence of points, not an image directly. To compare this sequence to the target image and, in turn, calculate the gradients w.r.t.~the generated points, we introduce a differentiable rendering module inspired by~\cite{qian2020end}. Third, we identify several challenges in optimization and propose corresponding solutions to make the overall learning process stable, effective, and efficient. The learned parameters can thus lead to high-quality fractal images that are visually similar to the target image. We note that besides reconstructing a target image, our approach can easily be repurposed by plugging in other loss functions that return gradients w.r.t.~the generated image.
 
We conduct two experiments to validate our approach. First, given a target image (that is not necessarily a fractal), we find the fractal parameters to reconstruct it and evaluate our approach using the pixel-wise mean square error (MSE). Our approach notably outperforms the baselines. Second, we extend our approach to finding a set of parameters to approximate a set of target images. This can be achieved by replacing the image-to-image MSE loss with set-to-set losses like the GAN loss~\cite{goodfellow2014generative}. In this experiment, we demonstrate the potential of finding more suitable fractal images for the downstream tasks than the randomly sampled ones~\cite{kataoka2020pre,Anderson_2022_WACV}.

\paragraph{Contributions.} We propose a gradient-descent-based approach to find the fractal parameters for a given image. Our approach is stable, effective, and efficient. It can be easily implemented using popular deep learning frameworks. Moreover, it can be easily extended from finding fractal parameters for image reconstruction to other purposes, by plugging the corresponding loss functions. While we mainly demonstrate the effectiveness of our approach using image reconstruction, we respect think it is a critical and meaningful step towards unlocking other potential usages and applicability of fractal images in visual recognition and deep learning. The codes are provided at \url{https://github.com/andytu28/LearningFractals}.

\section{Related Work}
\label{related}

\paragraph{Fractal inversion.} Finding the parameters of a given fractal image is a long-standing open problem~\cite{vrscay1991iterated,iacus2005approximating,kya2001optimization,guerin2005fractal,graham2019applying}. Many approaches are based on genetic algorithms~\cite{lankhorst1995iterated,gutierrez2000hybrid,nettleton1994evolutionary} or moment matching~\cite{vrscay1989iterated,forte1995solving}. Our approach is novel by tackling the problem via gradient descent and can be easily implemented using deep learning frameworks. We mainly focus on image reconstruction in the image/pixel space, while other ways of evaluation might be considered in the literature.

\paragraph{Fractal applications.}
Several recent works leveraged fractals to generate geometrically-meaningful synthetic data to facilitate neural network training~\cite{hendrycks2021pixmix,baradad2021learning,Ma_2021_ICCV,Anderson_2022_WACV,kataoka2020pre, gupta2021beyond, nakashima2021can}.
Some other works leveraged the self-similar property for image compression, for example, by partitioned IFS~\cite{fisher1994fractal}, which resembles the whole image with other parts of the same image \cite{pandey2015fractal, al2020review, jacquin1992image, al2016crowding, venkatasekhar2013fast, menassel2018improved,poli2022self}.
Fractals have also been used in other scientific fields such as diagnostic imaging
\cite{karperien2008automated}, 3D modeling
\cite{ullah2021utilizing}, optical systems~\cite{sweet1999topology}, biology~\cite{otaki2021fractal,enright2005mass}, etc. In this paper, we study the inverse problem, which has the potential to benefit these applications.

\paragraph{Fractals and neural networks.}
Fractals also inspire neural network (NN) architecture designs \cite{larsson2016fractalnet, deng2021fractal} and help the understanding of underlying mechanisms of NNs~\cite{camuto2021fractal,dym2020expression}. 
Specifically, some earlier works~\cite{kolen1994recurrent,tino1998recurrent,tino2004markovian} used the IFS formulation to understand the properties of recurrent neural networks (RNNs). These works are starkly different from ours in two aspects. First, we focus on the inverse problem of fractals. Second, we formulate an IFS as a special RNN to facilitate solving the inverse problem via gradient descent. We note that \citet{stark1991iterated} formulated an IFS as sparse feed-forward networks but not for solving the inverse problem.

\paragraph{Inversion of generative models.} The problem we studied is related to the inversion of generative models such as GANs~\cite{zhu2016generative,xia2021gan}. The goal is to find the latent representation of GANs for a given image that is not necessarily generated by GANs. Our problem has some unique challenges since the fractal generation process is stochastic and not directly end-to-end differentiable.
\section{Background}
\label{background}

\begin{figure*}[t]
\begin{minipage}{.46\linewidth}
\begin{algorithm}[H]
\footnotesize
\SetAlgoLined
\caption{IFS generation process.\\ See~\autoref{background} for details.}
\label{alg:ifs}
\SetKwInOut{Input}{Input}
\SetKwInOut{Output}{Output}
\Input{Fractal parameter $\sS$, \# of iterations $T$, an initial point $\vv^{(0)}\in\R^2$}
\For{$t\leftarrow 1$ \KwTo $T$}{
{\textbf{Sample} $(\mA^{(t)}, \vb^{(t)})$ from $\sS$}\\
{$\vv^{(t)} = \mA^{(t)} \vv^{(t-1)} +\vb^{(t)}$}
}
\Output{$\{\vv^{(0)},\cdots,\vv^{(T)}\}$}
\end{algorithm}
\end{minipage}
\hfill
\begin{minipage}{.46\linewidth}
\begin{algorithm}[H]
\footnotesize
\SetAlgoLined
\caption{IFS generation process via the \FE layer.\\ See~\autoref{ss_deep} for details.}
\label{alg:deep_ifs}
\SetKwInOut{Input}{Input}
\SetKwInOut{Output}{Output}
\Input{Fractal embedding $\{\mS_\text{A}, \mS_\text{b}\}$, \# of iterations $T$, an initial point $\vv^{(0)}\in\R^2$}
{\textbf{Sample} index sequence $\vz = [z^{(1)}, \cdots, z^{(t)}]$;}\\
\For{$t\leftarrow 1$ \KwTo $T$}{
{$\vv^{(t)}=\texttt{mat}(\mS_\text{A}[:, z^{(t)}]) \vv^{(t-1)} + \mS_\text{b}[:,
z^{(t)}]$;}
}
\Output{$\{\vv^{(0)},\cdots,\vv^{(T)}\}$}
\end{algorithm}
\end{minipage}
\vskip -15pt
\end{figure*}

We first provide the background about fractal generation via Iterated Function Systems (IFS)~\cite{barnsley2014fractals}. 
As briefly mentioned in~\autoref{intro}, an IFS can be thought of as ``drawing'' points iteratively on a canvas. The point transition is governed by a small set of $2\times 2$ affine transformations $\sS$: 
\begin{align}
    \label{eq:fractal_code}
	\sS = \left\{\left(\mA_n\in\R^{2\times 2}, \vb_n\in\R^2, p_n\in\R \right)\right\}_{n=1}^N. 
\end{align}
Here, an $(\mA_n, \vb_n)$ pair specifies an affine transformation:
\begin{align}
\label{eq:IFS}
\mA_n {\vv} +\vb_n,
\end{align}
which transforms a 2D position $\vv\in\R^2$ to a new position. The $p_n$ in \autoref{eq:fractal_code} is the corresponding probability of each transformation, \ie, $\sum_n p_n = 1$ and $p_n\geq 0, \forall n\in[N]$. $N$ is typically $2\sim10$. 

Given $\sS$, an IFS generates an image as follows. Starting from an initial point ${\color{red}\vv^{(0)}}\in\R^2$, an IFS repeatedly samples a transformation $(\mA^{(t)}, \vb^{(t)})$ from $\{(\mA_n, \vb_n )\}_{n=1}^N$ with replacement following the probability $\{p_n\}_{n=1}^N$, and applies 
\begin{align}
\label{eq:IFS}
{\color{red}\vv^{(t)}} = \mA^{(t)} {\color{red}\vv^{(t-1)}} +\vb^{(t)}
\end{align} 
to arrive at the next 2D point. This stochastic process can continue forever, but in practice we set a pre-defined number of iterations $T$.
The traveled points $\{\vv^{(0)},\cdots,\vv^{(T)}\}$ are then used to synthesize a fractal image. For example, one can quantize them into integer pixel coordinates and render each pixel as a binary (\ie, $1$) value on a black (\ie, $0$) canvas. We summarize the IFS fractal generation process in \autoref{alg:ifs}. Due to the randomness in sampling $(\mA^{(t)}, \vb^{(t)})$, one $\sS$ can create different but geometrically-similar fractal images. We call $\sS$ the fractal parameters.

\paragraph{Simplified parameters.} We follow~\cite{kataoka2020pre,Anderson_2022_WACV} to set $p_n$ by the determinant of the corresponding $\mA_n$. That is, $p_n\propto|\texttt{det}(\mA_n)|$. In the following, we will ignore $p_n$ when defining $\sS$:
\begin{align}
    \label{eq:fractal_code_simple}
	\sS = \left\{\left(\mA_n, \vb_n\right)\right\}_{n=1}^N. 
\end{align}

\section{End-to-End Learnable Fractals by Gradient Descent}
\label{difffractal}

In this paper, we study the problem: \emph{given a target image, can we find the parameters $\sS$ such that the generated IFS fractal image looks like it?}
Let us first provide the problem definition and list the challenges, followed by our algorithm and implementation.

\subsection{Problem and challenges}
\paragraph{Problem.} Given a gray-scaled image $\mI\in [0,1]^{H\times W}$, where $H$ is the height and $W$ is the width, we want to find the fractal parameters $\sS$ (cf.~\autoref{eq:fractal_code_simple}) such that the generated IFS fractal image $\mI'\in[0,1]^{H\times W}$ is close to $\mI$. We note that $\mI$ may not be a fractal image generated by an IFS. 

Let us denote the IFS generation process by $\mI'=G(\sS)$, and the image-to-image loss function by $\sL$. This problem can be formulated as an optimization problem:
\begin{align}
\min_{\sS} \sL(G(\sS), \mI). \label{eq_basic}
\end{align}
We mainly consider the pixel-wise square loss $\sL(\mI',\mI)=\|\mI'-\mI\|_2^2$, but other loss functions can be easily applied. \emph{We note that $G$ is stochastic, and we will discuss how to deal with it in \autoref{ss_opt}.}

\paragraph{Goal.} We aim to solve \autoref{eq_basic} via gradient descent (GD) --- as shown in~\autoref{eq:IFS}, each iteration in IFS is differentiable. Specifically, we aim to develop our algorithm as a module such that it can be easily implemented by modern deep learning frameworks like PyTorch and TensorFlow. This can largely promote our algorithm's usage, \eg, to combine with other modules for end-to-end training.

It is worth mentioning that while $\mI'$ depends on $\{p_n\}_{n=1}^N$ and $\{p_n\}_{n=1}^N$ depends on $\{\mA_n\}_{n=1}^N$ (cf. \autoref{eq:fractal_code_simple}), we do not include this path in deriving the gradients w.r.t.~$\{\mA_n\}_{n=1}^N$ for simplicity.

\paragraph{Challenges.} We identify several challenges in achieving our goal. 
\begin{itemize} [leftmargin=*,itemsep=0pt,topsep=0pt]
\item First, it is nontrivial to implement and parallelize the IFS generation process (for multiple fractals) using modern deep learning frameworks.
\item Second, the whole IFS generation process is not immediately differentiable. Specifically, an IFS generates a sequence of points, not directly an image. We need to back-propagate the gradients from an image to the points. 
\item Third, there are several technical details to make the optimization effective and stable. For example, an IFS is stochastic. The IFS generation involves hundreds or thousands of iterations (\ie, $T$); the back-propagation would likely suffer from exploding gradients.
\end{itemize} 

In the rest of this section, we describe our solution to each of these challenges. We provide a summarized pipeline of our approach in~\autoref{fig:render}.

\begin{figure*}[t]
    \centering
    \minipage{0.47\textwidth}
    \centering
    \includegraphics[width=1.\linewidth]{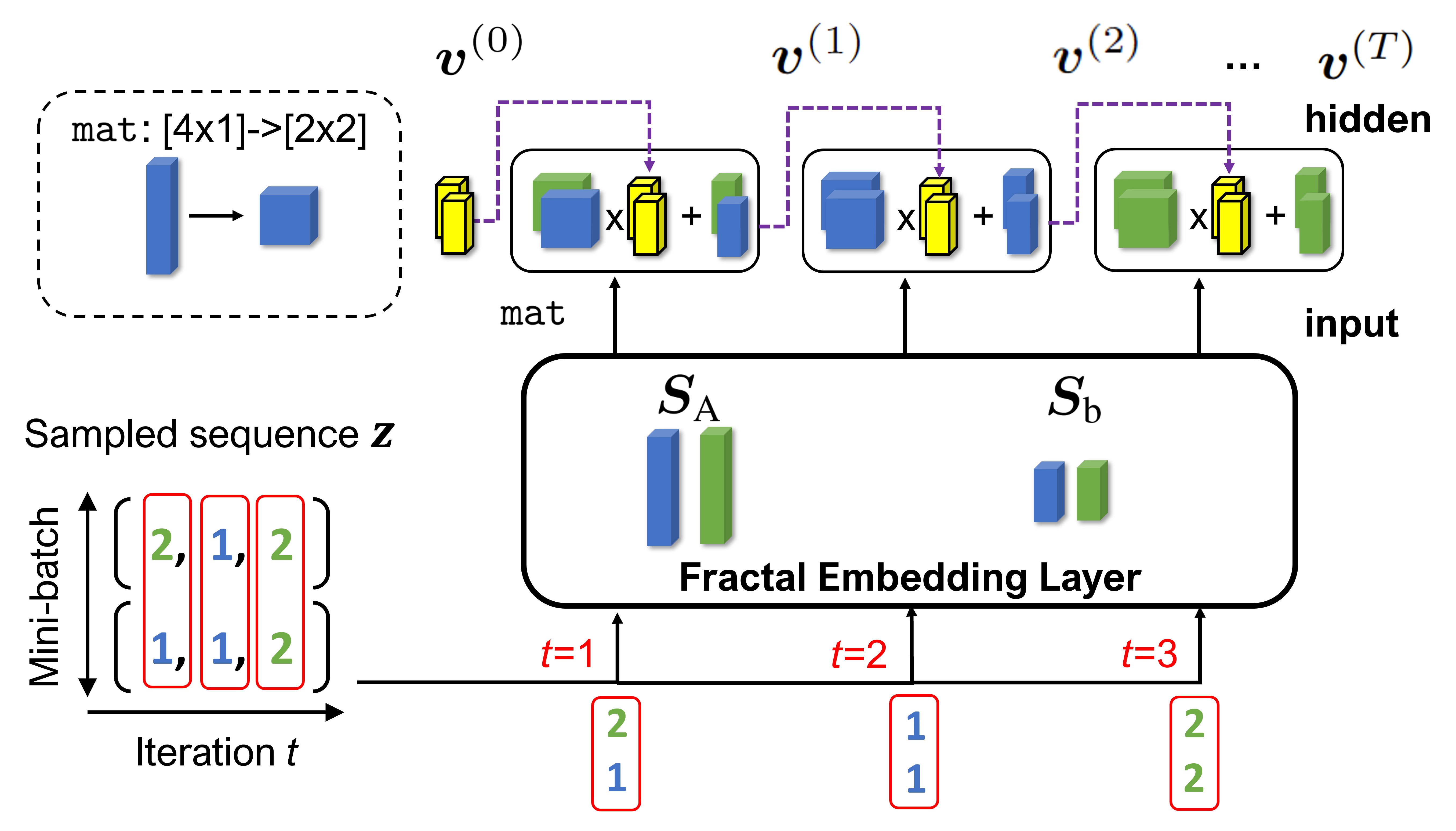} 
    \mbox{(a) }
    \endminipage
    \hfill
    \minipage{0.5\textwidth}
    \centering
    \includegraphics[width=1.\linewidth]{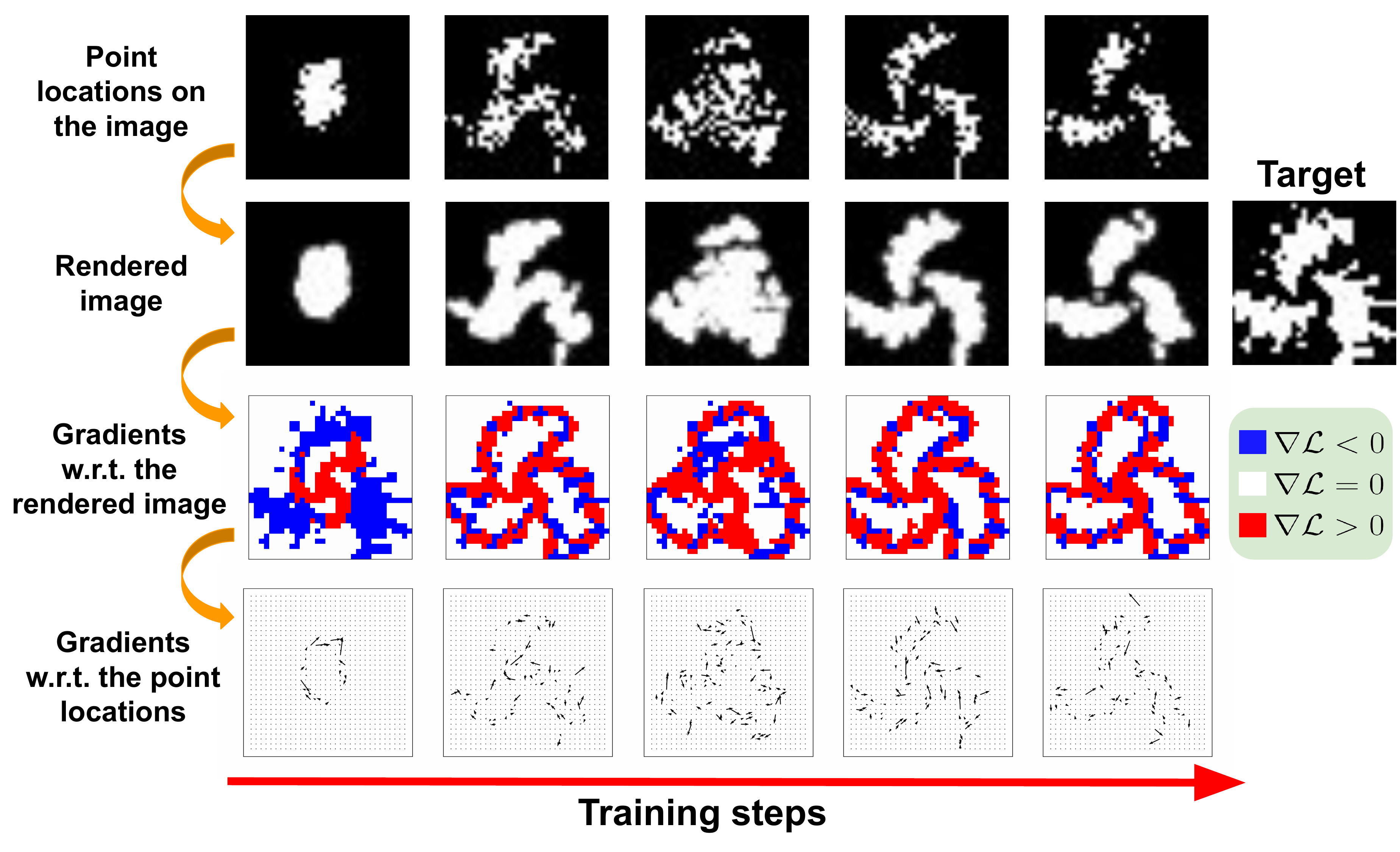}
    \vskip -6pt
    \mbox{(b)}
    \endminipage
    \vskip -6pt
    \caption{\small (a) \textbf{Fractal embedding layer} in~\autoref{ss_deep}. We show a case of $T=3$ and batch size $=2$. The purple dashed lines indicate the point transition (like the hidden vector in RNNs). (b) \textbf{Differentiable rendering} in~\autoref{ss_diff}. We show that the gradients w.r.t.~the rendered image can be translated into 2D gradient vectors w.r.t. the IFS points, \ie, $\nabla_{\vv^{(t)}}\sL\in\R^2, \forall t\in\{1,\cdots,T\}$ (zoom-in for better resolutions).}
    \vskip -10pt
    \label{fig:diff}
\end{figure*} 

\subsection{Fractal generation as a deep learning model}
\label{ss_deep}
\paragraph{IFS as an RNN.} We start with how to implement the IFS fractal generation process as a deep learning model. We note that an IFS performs \autoref{eq:IFS} for $T$ iterations. Every iteration involves an affine transformation, which is sampled with replacement from the fractal parameters $\sS$. Let us assume that $\sS$ contains only one pair $(\mA, \vb)$, then an IFS can essentially be seen as a special recurrent neural network (RNN), which has an identity activation function but no input vector:
\begin{align}
\vv^{(t)} = \mA\vv^{(t-1)} + \vb.
\end{align}
Here, $\vv^{(t)}$ is the RNN's hidden vector. That is, an IFS can be implemented like an RNN if it contains only one transformation. In the following, we extend this notion to IFS with multiple transformations. 

\paragraph{IFS rearrangement.} To begin with, let us rearrange the IFS generation process. Instead of sampling an affine transformation from $\sS$ at each of the $T$ IFS iterations, we choose to sample all $T$ of them before an IFS starts. We can do so because the sampling is independent of the intermediate results of an IFS.
Concretely, according to $\{p_n\}_{n=1}^N$, we first sample an index sequence $\vz = [z^{(1)}, \cdots, z^{(t)}]$, in which $z^{(t)}\in\{1, \cdots, N\}$ denotes the index of the sampled affine transformation from $\sS$ at iteration $t$.  
With the pre-sampled $\vz$, we can perform an IFS in a seemingly deterministic way. This is reminiscent of the re-parameterization trick commonly used in generative models~\cite{kingma2013auto}.

\paragraph{Fractal embedding layer.} We now introduce a \emph{fractal embedding} (\FE) layer, which leverages the rearrangement to implement an IFS like an RNN. The \FE layer has two sets of parameters, {\color{red}$\mS_\text{A}\in\R^{4\times N}$} and {\color{red}$\mS_\text{b}\in\R^{2\times N}$}, which record all the parameters in $\sS$ (cf., \autoref{eq:fractal_code_simple}). Specifically, the $n$\ts{th} column of $\mS_\text{A}$ (denoted as $\mS_\text{A}[:, n]$) records the $2\times 2$ parameters of $\mA_n$; \ie, $\mA_n=\texttt{mat}(\mS_\text{A}[:, n])$, where $\texttt{mat}$ is a reshaping operation from a column vector into a matrix. The $n$\ts{th} column of $\mS_\text{b}$ (denoted as $\mS_\text{b}[:,n]$) records $\vb_n$, \ie, $\vb_n = (\mS_\text{b}[:,n])$. Let us define one more notation $\textbf{1}_{z^{(t)}}\in\R^N$, which is an $N$-dimensional one-hot vector whose $z^{(t)}$\ts{th} element is $1$. With these parameters and notation, the \FE layer carries out the IFS computation at iteration $t$ (cf. \autoref{eq:IFS}) as follows
\begin{align}
\FE(z^{(t)}, \vv^{(t-1)}; \{\mS_\text{A}, \mS_\text{b}\})\nonumber
\end{align}
\begin{align}
&= \texttt{mat}(\mS_\text{A}\textbf{1}_{z^{(t)}}) \vv^{(t-1)} + \mS_\text{b}\textbf{1}_{z^{(t)}} \nonumber\\ 
&= \texttt{mat}(\mS_\text{A}[:, z^{(t)}]) \vv^{(t-1)} + \mS_\text{b}[:,
z^{(t)}] \\
&= \mA^{(t)} \vv^{(t-1)} +\vb^{(t)}. \nonumber
\end{align}
That is, the \FE layer takes the index of the  sampled affine transformation of iteration $t$ (\ie, $z^{(t)}$) and the previous point $\vv^{(t-1)}$ as input, and outputs the next point $\vv^{(t)}$. To perform an IFS for $T$ iterations, we only need to call the \FE layer recurrently for $T$ times. This analogy between IFS and RNNs makes it fairly easy to implement the IFS generation process in modern deep learning frameworks. Please see~\autoref{alg:deep_ifs} for a summary.  

\paragraph{Extension to mini-batch versions.} The \FE layer can easily be extended into a mini-batch version to generate multiple fractal images, if they share the same fractal parameters $\sS$. See~\autoref{fig:diff} (a) for an illustration. 
Interestingly, even if the fractal images are based on different fractal parameters, a simple parameter concatenation trick can enable mini-batch computation.
Let us consider a mini-batch of size $2$ with fractal parameters $\sS_1$ and $\sS_2$. We can combine $\sS_1$ and $\sS_2$ into one $\mS_\text{A}\in\R^{4\times (2N)}$ and 
one $\mS_\text{b}\in\R^{2\times (2N)}$, where the number of columns in $\mS_\text{A}$ and $\mS_\text{b}$ are doubled to incorporate the affine transformations in $\sS_1$ and $\sS_2$. When sampling the index sequence for each fractal, we just need to make sure that the indices align with the columns of $\mS_\text{A}$ and $\mS_\text{b}$ to generate the correct fractal images.

\subsection{Differentiable rendering}
\label{ss_diff}
The \FE layer enables us to implement an IFS via modern deep learning frameworks, upon which we can enjoy automatic back-propagation to calculate the gradients $\nabla_{\sS}\sL$ (cf. \autoref{eq_basic}) if we have obtained the gradients $\nabla_{\vv^{(t)}}\sL, \forall t\in\{1,\cdots,T\}$. However, obtaining the latter is not trivial. An IFS generates a sequence $\{\vv^{(0)},\cdots,\vv^{(T)}\}$ of 2D points, but it is not immediately clear how to render an image $\mI'$ such that subsequently we can back-propagate the gradient $\nabla_{\mI'}\sL(\mI', \mI)$ to obtain $\nabla_{\vv^{(t)}}\sL$. 

Concretely, $\nabla_{\mI'}\sL$ is a 2D tensor of the same size as $\mI'$, where $\nabla_{\mI'[h, w]}\sL\in\R$ indicates if the pixel $[h, w]$ of $\mI'$ should get brighter or darker. Let us follow the definition in~\autoref{background}: IFS draws points on a black canvas. Then if $\nabla_{\mI'[h, w]}\sL\in\R<0$, ideally we want to have some IFS points closer to or located at $[h, w]$, to make $\mI'[h, w]$ brighter. To do so, however, requires a translation of $\nabla_{\mI'[h, w]}\sL<0$ into a ``pulling'' force towards $\mI'[h, w]$ for some of the points in $\{\vv^{(0)},\cdots,\vv^{(T)}\}$ .

To this end, we propose to draw the IFS points $\{\vv^{(0)},\cdots,\vv^{(T)}\}$\footnote{Following~\cite{kataoka2020pre,Anderson_2022_WACV}, we linearly translate and re-scale the points to fit them into the $H\times W$ image plane.} onto $\mI'$ using an RBF kernel --- every point ``diffuses'' its mass to nearby pixels. This is inspired by the differentiable soft quantization module~\cite{qian2020end} proposed for 3D object detection. The rendered image $I'$ is defined as 
\begin{align}
\label{eq:render}
\mI'[h, w] = \sum_{t} \exp(-\left\|
[h, w]^\top
-\vv^{(t)}\right\|_2^2 / \tau), \nonumber \\ 
h \in [H], w \in [W],
\end{align}
where $\tau>0$ governs the kernel's bandwidth. This module not only is differentiable w.r.t.~$\vv^{(t)}$, but also fulfills our requirement. The closer the IFS points to $[h, w]$ are, the larger the intensity of $\mI'[h, w]$ is. Moreover, if $\nabla_{\mI'[h, w]}\sL<0$, then $\frac{\partial \sL}{\partial \mI'[h, w]}\times \frac{\partial \mI'[h, w]}{\partial \vv^{(t)}}$ will be a 2D vector positively proportional to $(\vv^{(t)} - [h, w]^\top)$. When GD is applied, this will pull $\vv^{(t)}$ towards $[h, w]^\top$. See \autoref{fig:diff} (b) for an illustration.

Since $\mI'[h, w]$ may have a value larger than $1$, we clamp it back to the range $[0, 1]$ by $\min(\mI'[h, w], 1)$.

\subsection{Objective and optimization}
\label{ss_opt}
In \autoref{ss_deep} and \autoref{ss_diff}, we present an end-to-end differentiable framework for learning $\sS$. In this subsection, we discuss the objective function and present important technical insights to stabilize the optimization.

\paragraph{Objective function.} As pointed out in the subsequent paragraph of \autoref{eq_basic}, since an IFS is stochastic, the objective function in \autoref{eq_basic} is ill-posed. To tackle this, let us rewrite the stochastic $G(\sS)$ by $G(\sS; \vz)$, in which we explicitly represent the stochasticity in $G(\sS)$ by $\vz$: the sampled index sequence according to $\sS$ (see \autoref{ss_deep}). We then present two objective functions:
\begin{align}
\paragraph{Expectation: } \expect{\vz\sim\sS}{\sL(G(\sS, \vz), \mI)},\nonumber \\ \textbf{Fixed: } \expect{\vz\in\sZ}{\sL(G(\sS, \vz), \mI)}, \label{eq_new_obj}
\end{align}
where $\vz\sim\sS$ indicates that $\vz$ is sampled according to $\sS$; $\sZ$ is a set of pre-sampled fixed sequences. For instance, we can sample $\sZ$ according to the initial $\sS$ and fix it throughout the optimization process. The {Expectation} objective encourages every $\vz\sim\sS$ to generate $\mI$. The learned $\sS$ thus would easily generate images $\mI'$ close to $\mI$, but the diversity among $\mI'$ might be limited. The {Fixed} objective is designed to alleviate this limitation by only forcing a (small) set of fixed sequences $\sZ$ to generate $\mI$. However, since $\sZ$ is pre-sampled, their probability of being sampled from the learned $\sS$ might be low. Namely, it might be hard for the learned $\sS$ to generate images like $\mI$. We evaluate both in~\autoref{exp}.

\paragraph{Optimization.} We apply mini-batch stochastic gradient descent (SGD) to optimize the objectives w.r.t.~$\sS$ in~\autoref{eq_new_obj}. 
To avoid confusion, we use ``iteration'' for the IFS generation process, and ``step'' for the SGD optimization process. The term ``stochastic'' refers to sampling a mini-batch of $\vz$ at every optimization step from either $\sS$ or $\sZ$, generating the corresponding fractal images $\mI'$ by IFS, and calculating the average gradients w.r.t.~$\sS$.

\begin{figure}[t]
\centering
\includegraphics[width=0.7\linewidth]{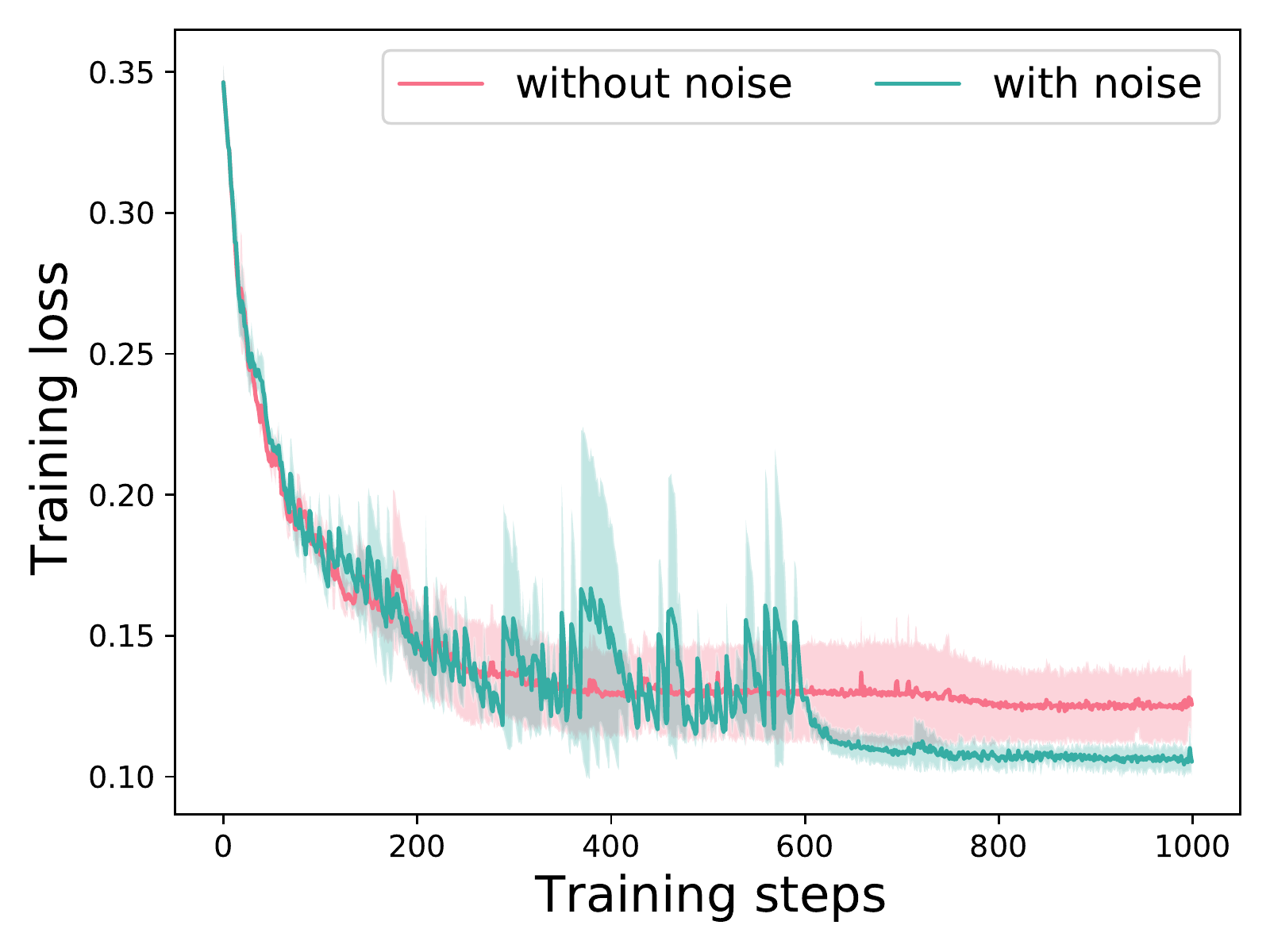}
\vskip -5pt
\caption{\textbf{The training loss along the steps of SGD w/ and w/o adding random noise to the fractal parameters $\sS$.} We add a Gaussian noise $\sim\mathcal{N}(0, 0.1)$ to each element of the fractal parameters $\sS$ every $5$ SGD steps. This makes the final loss lower and the optimization robust. (The ``shaded'' area is the standard deviation across different random seeds.)}   
\vskip -10pt
\label{fig:stable}
\end{figure}

We note that unlike neural networks that are mostly over-parameterized, $\sS$ has much fewer, usually tens of parameters. This somehow makes the optimization harder. Specifically, we found that the optimization easily gets stuck at poor local minima. (In contrast, over-parameterized models are more resistant to it~\cite{du2018gradient,allen2019learning,arora2019fine}.) To resolve this, whenever we calculate the stochastic gradients, we add a small Gaussian noise to help the current parameters escape local minima, inspired by~\cite{welling2011bayesian}. \autoref{fig:stable} shows that this design stabilizes the optimization process and achieves lower loss. More results are in~\autoref{exp}.

\paragraph{Reparameterized $\sS$.} 
In the extreme case where $\sS$ contains only one affine transformation $(\mA, \vb)$, recurrently applying it multiple times
could easily result in exploding gradients when the maximum singular value of $\mA$ is larger than $1$~\cite{pascanu2013difficulty}. 
This problem happens when $\sS$ contains multiple affine transformations as well, if most of the matrices have such a property. 
To avoid it, we follow~\cite{Anderson_2022_WACV} to decompose ``each'' $\mA$ in $\sS$ into rotation, scaling, and flipping matrices
\begin{align}
    & {\tiny
    \mA=\underbrace{\begin{bmatrix}
    \text{cos}(\theta) & -\text{sin}(\theta) \\
    \text{sin}(\theta) & \text{cos}(\theta) 
    \end{bmatrix}}_{\text{rotation}} 
    \underbrace{\begin{bmatrix}
    \sigma_1 & 0 \\
    0 & \sigma_2
    \end{bmatrix}}_{\text{scaling}} 
    \underbrace{\begin{bmatrix}
    \text{cos}(\phi) & -\text{sin}(\phi) \\
    \text{sin}(\phi) & \text{cos}(\phi) 
    \end{bmatrix}}_{\text{rotation}} 
    \underbrace{\begin{bmatrix}
    d_1 & 0 \\
    0 & d_1
    \end{bmatrix}}_{\text{flipping}},
    } \label{eq_repara}
\end{align}
where $\sigma_1, \sigma_2 \geq 0; d_1, d_2 \in \{-1, 1\}$. That is, $\mA$ is \emph{reparameterized} by $\{\theta, \phi, \sigma_1, \sigma_2, d_1, d_2\}$.
We note that \citet{Anderson_2022_WACV} used this decomposition to sample $\mA$. The purpose was to generate diverse fractal images for pre-training.
Here, we use this decomposition for a drastically different purpose: to stabilize the optimization of $\mA$. We leverage one important property --- $\max(\sigma_1, \sigma_2)$ is exactly the maximum singular value of $\mA$ --- and clamp both $\sigma_1$ and $\sigma_2$ into $[0, 1]$ to avoid exploding gradients.

\begin{figure*}[t!]
\centering
\includegraphics[width=0.75\textwidth]{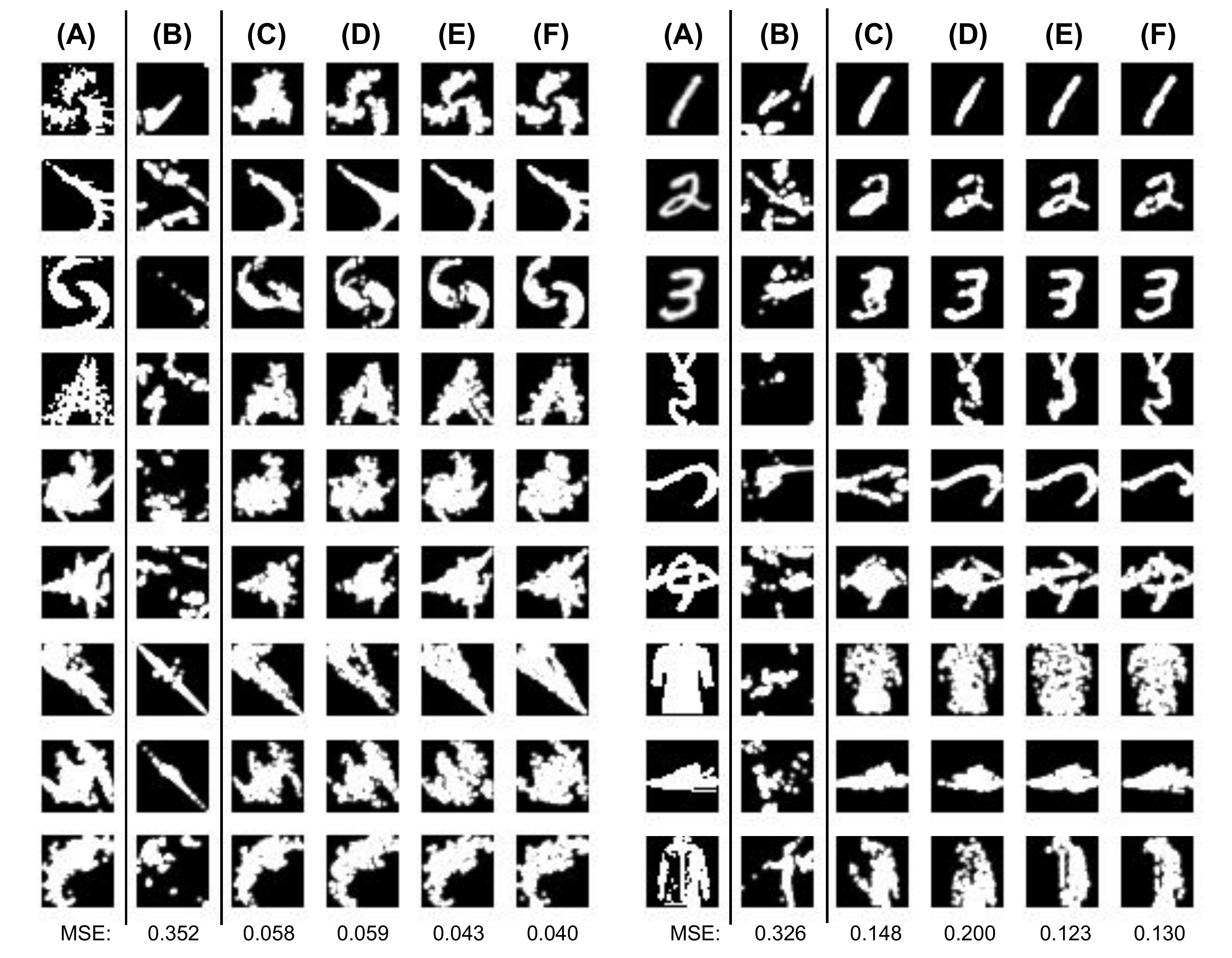}
\vskip -10pt
\caption{\small Image reconstruction examples. Left: FractalDB images. Right: MNIST (row 1-3), KMNIST (row 4-6), and FMNIST images (row 7-9). (A): the target images. (B): an inverse encoder baseline. (C-F):  variants of our approach (corresponding to the ``Index'' column in~\autoref{tbl:inversion_mse}).  The last row shows two examples and the corresponding MSE.}   
\label{fig:inversion}
\end{figure*}

\begin{table*}[h!]
\footnotesize
\centering
\setlength{\tabcolsep}{1.5pt} 
\vskip -0pt          
\caption{\small Averaged MSE (mean$\pm$std over $3$ runs) of image reconstruction on FractalDB/MNIST/KMNIST/FMNIST dataset.}
\vskip -7pt
\begin{tabular}{l|c|ccc|cccc}
\toprule
 Index & Method & Objective & Clamp & Noisy & FractalDB & MNIST & KMNIST & FMNIST \\
\midrule
(B) & Inverse encoder & - & - & - & $0.3889 \pm 0.0000$ & $0.1724 \pm 0.0000$ & $0.3016 \pm 0.0000$ & $0.3658 \pm 0.0000$ \\
\midrule
(C) &\multirow{4}{*}{Ours} & Fixed & \xmark & \xmark & $0.0790\pm0.0006$ & $0.0475 \pm 0.0012$ & $0.1023 \pm 0.0016$ & $0.0692 \pm 0.0010$ \\
(D) & &Fixed & \cmark & \xmark & $0.0752\pm0.0027$ & $0.0312 \pm 0.0002$ & $0.0979 \pm 0.0008$ & $0.0765 \pm 0.0034$ \\
(E) & &Expectation & \cmark & \xmark & $0.0630\pm0.0008$ & $0.0223 \pm 0.0004$ & $0.0826 \pm 0.0010$ & $0.0547 \pm 0.0005$ \\
(F) & &Expectation & \cmark & \cmark & $\mathbf{0.0617\pm0.0009}$ & $\mathbf{0.0202 \pm 0.0005}$ & $\mathbf{0.0781 \pm 0.0015}$ & $\mathbf{0.0538 \pm 0.0006}$ \\
\midrule
& Ours (w/o Reparam. $\sS$) & Expectation & \cmark & \cmark  & Exploded & Exploded & Exploded & Exploded\\
\bottomrule
\end{tabular}
\vskip -5pt
\label{tbl:inversion_mse}
\end{table*}

\subsection{Extension}
\label{ss_extension}
Besides the pixel-wise square loss $\sL(\mI',\mI)=\|\mI'-\mI\|_2^2$, our approach can easily be compatible with other loss functions on $\mI'$. The loss may even come from a downstream task. In~\autoref{ss_GAN}, we experiment with one such extension. Given a set of images $\{\mI_m\}_{m=1}^M$, we aim to learn a set of fractal parameters $\{\sS_j\}_{j=1}^J$ such that their generated images are distributionally similar to $\{\mI_m\}_{m=1}^M$. We apply a GAN loss~\cite{goodfellow2014generative}. In the optimization process, we train a discriminator to tell real images from generated images, and use it to provide losses to the generated images.

\section{Experiments}
\label{exp} 

We conduct two experiments to quantitatively and qualitatively validate our approach and design choices. These include (1) \textbf{image reconstruction:} learning an IFS fractal to reconstruct each given target image by gradient descent, using the mean squared error on pixels; (2) \textbf{learning fractals with a GAN loss:} given a set of target images, extending our approach to learn a set of fractal parameters that generates images with distributions similar to the targets. 

We note that \emph{we do not intend to compete with other state-of-the-art in these tasks that do not involve fractals}. Our experiments are to demonstrate that our learnable fractals can capture meaningful information in the target images. \emph{Please see the supplementary material for more details and analyses.}

\subsection{Setups}
\paragraph{Datasets.}
We first reconstruct random fractal images generated following \textbf{FractalDB}~\cite{kataoka2020pre,Anderson_2022_WACV}. 
We then consider images that are not generated by fractals, including \textbf{MNIST} (hand-written digits)~\cite{lecun1998gradient}, \textbf{FMNIST} (fashion clothing)~\cite{Xiao2017FashionMNISTAN}, and \textbf{KMNIST} (hand-written characters)~\cite{clanuwat2018deep}, for both reconstruction and GAN loss experiments. 
We use $32\times32$ binarized (black-and-white) images as target images for reconstruction. 
For the GAN loss experiment, we use the $60K/10K$ training/test images out of $10$ classes in the three non-fractal datasets.

\paragraph{Technical details about fractals.}
Throughout the experiments, we sample $\vv^{(0)}$ and initialize $\sS$ by the random procedure in~\cite{Anderson_2022_WACV}. 
The number of transformations in $\sS$ is fixed to $N=10$, and we set $T=300$ per image. 
These apply to the FractalDB generation as well.  

To learn $\sS$ using our proposed method, we set $\tau=1$ (RBF kernel bandwidth) and apply an Adam optimizer~\cite{kingma2014adam}. We verify our design choices in the ablation studies of~\autoref{ss_inversion} and the supplementary material.

\subsection{Image reconstruction and ablation study}
\label{ss_inversion}

In the image reconstruction task, our goal is to verify our gradient descent-based method can effectively recover target shapes. 
Specifically, given a target image, we learn an $\sS$ using our proposed approach to reconstruct the target shape by minimizing the \textbf{mean squared error (MSE)} between the generated and the target images. 
Each $\sS$ is learned with a batch size $50$ and a learning rate $0.05$ for $1,000$ SGD steps. As we recover any given target by learning a fractal $\sS$, no training and test splits are considered.  
In this experiment, we generate $2K$ target images using FractalDB, and randomly sample $2K$ target images from each MNIST/FMNIST/KMNIST.
For quantitative evaluation, after $\sS$ is learned, we generate $100$ images per $\sS$ (with different sampled sequences $\vz$) and report the minimum MSE to reduce the influence of random sequences. We consider the following variants of our method:
\begin{itemize}[leftmargin=*,itemsep=0pt,topsep=0pt]
\item Using either the \emph{Expectation} objective or the \emph{Fixed} objective discussed in~\autoref{ss_opt}.
\item In differentiable rendering (see  \autoref{eq:render}), \emph{clamping} each pixel to be within $[0, 1]$ or not. 
\item Applying the \emph{noisy} SGD trick or not (see~\autoref{ss_opt}). 
\item Learning the reparameterized $\sS$ or the original IFS parameters $\sS$ (see~\autoref{ss_opt}).
\end{itemize}
Besides our variants, we also include an \textbf{inverse encoder} baseline similar to the recent work~\cite{graham2019applying} that learns a ResNet18~\cite{he2016deep} to predict the fractal parameters $\sS$ given an image (\ie, as a regression problem). We train such a regressor on $20K$ different FractalDB images (using the corresponding $\sS$ as labels) and use it on the target images (details are in the supplementary material).

Qualitatively, \autoref{fig:inversion} shows examples of image reconstruction. Our approach works well on fractal shapes. 
Surprisingly, we can also recover non-fractal datasets to an extent.
This reveals that IFS can be more expressive if the parameters are learned in a proper way. 
By contrast, the inverse encoder cannot recover the target shapes and ends up with poor MSE. 

Quantitatively, the reported MSE (averaged over all targets) in~\autoref{tbl:inversion_mse} verifies our design choices of applying the Expectation objective, clamping the rendered images, and applying noisy SGD. 
Notably, without reparameterization, we can hardly learn $\sS$ since the gradients quickly explode.

\begin{figure}[t]
\centering
\minipage{0.33\columnwidth}
\centering
\includegraphics[width=0.9\linewidth]{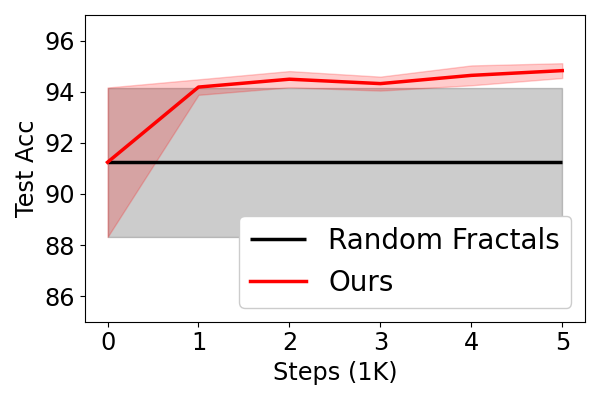}
\mbox{\small (a) MNIST}
\endminipage
\hfill
\minipage{0.33\columnwidth}
\centering
\includegraphics[width=0.9\linewidth]{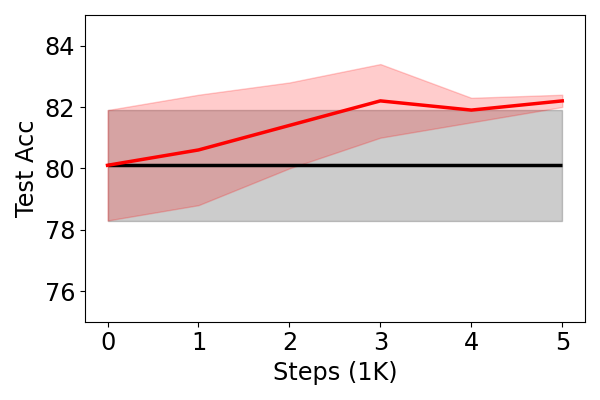}
\mbox{\small (b) FMNIST}
\endminipage
\hfill
\minipage{0.33\columnwidth}
\centering
\includegraphics[width=0.9\linewidth]{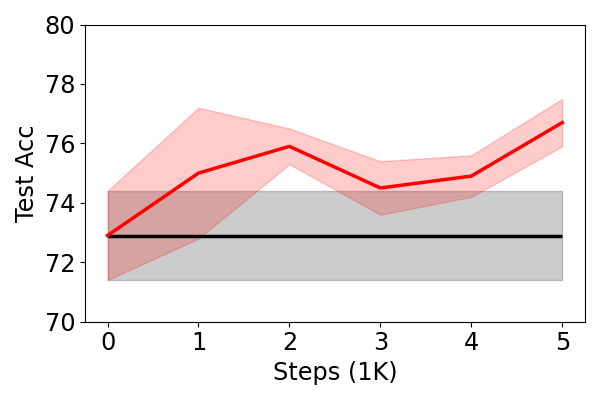}
\mbox{\small (b) KMNIST}
\endminipage
\hfill
\vskip -5pt
\caption{\small \textbf{Learning fractals as GANs.} We conduct linear evaluations using features pre-trained (for $0\sim5K$ SGD steps) on the {\color{red}learned}/{\color{gray}random} fractal images. Test accuracy of means (lines) and standard deviations (shades) over $10$ runs are plotted.}  
\label{fig:gan}
\vskip -5pt
\end{figure}

\subsection{Learning fractals as GAN generators}
\label{ss_GAN}

As discussed in~\autoref{ss_extension}, our approach is a flexible deep learning module that can be learned with other losses besides MSE.
To demonstrate such flexibility, we learn fractals like GANs~\cite{goodfellow2014generative} by formulating a set-to-set learning problem --- matching the distributions of the images generated from fractals to the distributions of (real) target images. 
Specifically, we start from $100$ randomly sampled fractals $\sS$ and train them on the target training set of MNIST/FMNIST/KMINIST in an unsupervised way using the GAN loss with a discriminator (a binary LeNet~\cite{lecun1998gradient} classifier). 
Next, we evaluate the learned fractals by using them for model pre-training~\cite{kataoka2020pre}. 

We argue that if the learned fractals $\sS$ can produce image distributions closer to the targets, pre-training on the generated images should provide good features w.r.t. the (real) target images.
We expect pre-training on the images generated from the learned $\sS$ to yield better features for the target datasets, compared to pre-training on random fractals.

Concretely, we pre-train a $100$-way classifier (using the LeNet architecture again) from scratch. We generate $600$ images from each $\sS$ and treat each $\sS$ as a pseudo-class following~\cite{kataoka2020pre} to form a $100$-class classification problem. Then, we conduct a linear evaluation on the pre-trained features using the real training and test sets. 

\autoref{fig:gan} shows the results, in which we compare pre-training on the random fractals (gray lines) and on the learned fractals (red lines), after every $1K$ SGD steps. 
As we pre-train $\sS$ with more steps, the learned fractals can more accurately capture the target image distribution, evidenced by the better pre-trained features that lead to higher test accuracy.  
This study shows the helpfulness of our method for pre-training.

\section{Conclusion, Limitations, and Future Work}
\label{con}

\begin{figure}
\centering
\vskip -5pt
\includegraphics[width=0.30\linewidth]{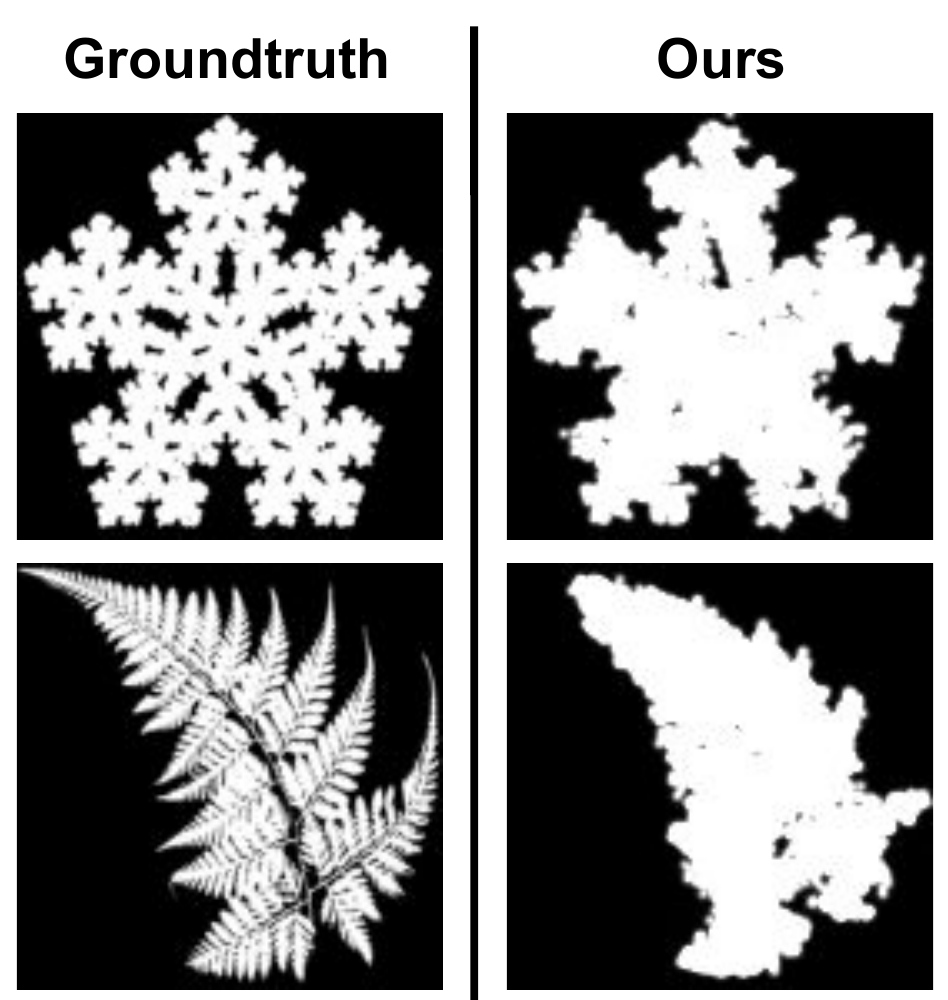}
\vskip -5pt
\caption{\small Inversion on complex (fractal) images.}  
\label{fig:real}
\vskip -15pt
\end{figure}

We present a gradient descent-based method to find the fractal parameters for a given image. 
Our approach is stable, effective, and can be easily implemented using popular deep learning frameworks. 
It can also be extended to finding fractal parameters for other purposes, \eg, to facilitate downstream vision tasks, by plugging in the corresponding losses.

We see some limitations yet to be resolved. For example, IFS itself cannot handle color images. Our current approach is hard to recover very detailed structures (\eg,~\autoref{fig:real}). We leave them for our future work. 
We also plan to further investigate our approach on 3D point clouds.

\section*{Acknowledgments}
This research is supported in part by grants from the National
Science Foundation (IIS-2107077, OAC-2118240, and OAC-2112606), the OSU GI Development funds, and Cisco Systems, Inc. We are thankful for the generous support of the computational resources by the Ohio Supercomputer Center and AWS Cloud Credits for Research. We thank all the feedback from the review committee and have incorporated them. 

{\small
\bibliography{main}
}

\clearpage
\newpage
\appendix
\section*{\LARGE Supplementary Material}

We provide details omitted in the main paper. 

\begin{itemize}
    \item \autoref{suppl-sec:train_details}: additional training details (cf. \autoref{exp} of the main paper).
    \item \autoref{suppl-sec:additional_exp}: additional experiments and analyses (cf. \autoref{exp} of the main paper). 
    \item \autoref{suppl-sec:additional_discussions}: additional discussions. 
\end{itemize}

\section{Additional Training Details} \label{suppl-sec:train_details}

\subsection{Training details for image inversion}
\emph{We note that for the image inversion experiment, our goal is to learn $\sS$ for a given target image. The quantitative error is directly measured between the generated images by the learned $\sS$ and the target image.}

We provide training details for the inverse encoder baseline~\cite{graham2019applying} we compared to that predicts fractal parameters $\sS$ (\ie, as a regression problem) given an input image (cf. \autoref{ss_inversion} of the main paper). We first pre-generate $100K$ fractal images from $20K$ different $\sS$; each $\sS$ has 10 transformations. For each image, its corresponding $\sS$ is used as its label (\ie, regression target), and the transformations are ordered by their maximum singular values. Then, we train a ResNet18 to directly output the target $\sS$, which has $10\times(4+2)$ dimensions, using the mean squared error (MSE). The ResNet18 is pre-trained on ImageNet, and we train it on fractal inversion for 100 epochs using the SGD optimizer with learning rate $0.05$, momentum $0.9$, weight decay $1\mathrm{e}{-4}$, and batch size $128$. The learning rate is reduced every epoch following the cosine schedule~\cite{loshchilov2017sgdr}.

The training details of our learnable fractals are as follows. Given a target image, we randomly initialize an $\sS$ in its reparameterized form (cf. \autoref{eq_repara}). Then, in all experiments, we use the Adam optimizer with a mini-batch of $50$ sequences (\ie, $\vz$) to train $\sS$. For the Fixed objective, we randomly sample the $50$ sequences based on the initial $\sS$ at the beginning and fix them throughout the training. For the Expectation objective, we sample different $50$ sequences at every training step. Our sequence sampling follows~\cite{Anderson_2022_WACV}. We optimize the $\sS$ for $1000$ steps, and the learning rate is initialized with $0.05$ and decayed by $0.5$ every $250$ steps.

\subsection{Training details for learning fractals as GAN}

\emph{We note that for this experiment, our goal is to learn a set of $\sS$ such that the resulting fractal images can be used to pre-train a LeNet with stronger features. We use supervised pre-training: images generated by each $\sS$ are considered as the same class.}

We start by randomly sampling 100 different fractals $\sS$ (each can be used to generate more fractal images) and gradually adjust them so that the overall distributions of their generated fractal images become closer to the real images in MNIST/FMNIST/KMNIST. 
To achieve this goal, we plug in the GAN loss to train the 100 fractal $\sS$ as a generator. That is, we randomly sample some $\sS$ and use them to generate a mini-batch of images. A discriminator network will compare the distribution of the generated images and the real images and compute the GAN loss for updating the generator. In the GAN training, we adopt batch size $16$ and generator learning rate $1\mathrm{e}{-4}$. We use a LeNet as our discriminator and update it once every $10$ training steps of the generator. The discriminator learning rate is $1\mathrm{e}{-5}$. 

As we gradually train the $100$ fractals $\sS$ to generate images similar to the real data, we expect the generated fractal images to become similar to the real image distributions. We note that this process is unsupervised as GAN training does not require any categorical labels from these datasets. 

We adopt a linear probing way to evaluate the quality of the generated images --- the generated images should be suitable to pre-train features for MNIST/FMNIST/KMNIST. We evaluate the pre-training performance as we keep updating the $\sS$. At every 1K steps of training the 100 fractal $\sS$, we use them to generate $60K$ images for pre-training LeNet features. The features are then evaluated by learning a linear classifier (with frozen features) on the real training sets of MNIST/FMNIST/KMNIST. In the linear evaluation, we train the classifier for 100 epochs using SGD with a learning rate $0.001$, momentum $0.9$, and a batch size $128$.

\section{Additional Experiments and Analyses} \label{suppl-sec:additional_exp}

\begin{table}[t]
\footnotesize 
\centering
\setlength{\tabcolsep}{3pt}
\vskip -0pt
\caption{Averaged MSE (means$\pm$std over $3$ runs) of our fractal inversion with $N\in{2, 6, 10}$ transformations.}
\begin{tabular}{c|ccc}
\toprule
Dataset & $N=2$ & $N=6$ & $N=10$ \\
\midrule
FractalDB & $0.0954\pm0.0035$ & $0.0687\pm0.0006$ & $0.0627\pm0.0013$ \\ 
MNIST     & $0.0423\pm0.0013$ & $0.0237\pm0.0015$ & $0.0212\pm0.0009$ \\
\bottomrule
\end{tabular}
\label{tbl:diff_num_trans}
\end{table}

\subsection{Different numbers of transformations in $\sS$}

As the number of transformations controls the expressibility of $\sS$, we provide an ablation study to investigate its influence on image inversion. We randomly sample $500$ images for FractalDB/MNIST and adopt our best setup (with clamping, the Expectation objective, and noisy SGD) for training with $N=2, 6, 10$. The averaged MSE is reported in~\autoref{tbl:diff_num_trans}. The results show that the MSE decreases as $N$ increases, reflecting how $N$ affects the expressive power of $\sS$. We conclude that target shapes can be more accurately recovered if more fractal parameters (\ie, $\sS$ with more affine transformations) are used.

\begin{table}[t]
\footnotesize 
\centering
\setlength{\tabcolsep}{3pt}
\vskip -0pt
\caption{Averaged MSE (means$\pm$std over $3$ runs) of our fractal inversion with Adam or SGD.}
\begin{tabular}{c|cc}
\toprule
Dataset & SGD & Adam \\
\midrule
MNIST     & $0.0656\pm0.0014$ & $0.0212\pm0.0009$ \\
\bottomrule
\end{tabular}
\label{tbl:sgd_vs_adam}
\end{table}

\subsection{Different optimizers (SGD vs. Adam)}

The optimization of fractals is non-trivial and involves several technical details to make it effective and stable. Throughout all our experiments, we adopt the Adam optimizer as we find it particularly effective for learning fractals. We provide the results of our best setup using the SGD optimizer with a learning rate $0.05$ and momentum $0.9$ to justify our choice. We conduct the comparison on randomly sampled $500$ MNIST images. Both the SGD and Adam optimizers are used to train fractals for $1000$ steps with a learning rate decayed by $0.5$ every $250$ steps. As shown in~\autoref{tbl:sgd_vs_adam}, the Adam optimizer obtains lower MSE than SGD with momentum, yielding better reconstruction performance.

\section{Additional Discussions} \label{suppl-sec:additional_discussions}

\subsection{More discussions on learning fractals by gradient descent}
To validate our approach for learning fractals, we conduct two tasks, including (1) image inversion: learning an IFS fractal to reconstruct each given image using gradient descent, along with mean squared error on pixels; (2) learning fractals with a GAN loss: given a set of images, extending our approach to learn a set of fractal parameters for them. We provide more discussions on them. 

We first study image inversion. Our objective is to verify our gradient-descent-based method can recover target shapes. In this study, we reconstruct any given target image by training a fractal $\sS$, so no training and test splits are considered. Several ablation studies are conducted to qualitatively and quantitatively validate our design choices. We find that using the Expectation objective gives the most significant gain, and the noisy SGD can further improve the inversion quality. The combination of these two, along with clamping pixels to be $[0, 1]$, yields the best setup of our method. 

Next, we adopt our best setup in GAN learning to demonstrate the benefits of making fractals learnable by gradient descent. \emph{We emphasize that our objective is not to obtain state-of-the-art accuracy but to show the potential of our learnable fractals that can capture geometrically-meaningful, discriminative information in the target images.} 
Specifically, we exploit the flexibility of optimizing via gradient descent to plug in another loss function, the GAN loss, to formulate a set-to-set learning. By gradually modifying random fractals~\cite{kataoka2020pre} to generate images similar to the target data, we can pre-train better features that are more suitable for downstream datasets (MNIST/FMNIST/KMNIST). For each dataset, we use its training set (without labels) to adjust $100$ random fractal $\sS$. Then, the features pre-trained using the adjusted fractals are evaluated following the standard linear evaluation.

\subsection{Potential negative societal impact} 
We provide an optimization framework for learning fractals by gradient descent. In the field of image generation, although we show that fractals can recover various digits and characters, their expressibility is highly constrained by the iterated function systems (IFS), making them less probable to create real-looking fake images like GANs. In deep learning, the main application of fractals is model pre-training. Our method is likely to be applied to improve upon those related studies when limited downstream data are accessible. Therefore, we share the negative societal impact (but less) with the field of image generation and pre-training to have a bias toward the collected real data. Nevertheless, we emphasize that this potential negative impact does not result from our algorithm.

\subsection{Computation resources}

We conduct our experiments on PyTorch and on $4$ NVIDIA RTX A6000 GPUs. It takes roughly $2$ minutes to learn a fractal $\sS$ for $1$ target image on $1$ GPU. Throughout the paper, we train on roughly $103.5K$ images of different datasets and settings, thereby resulting in a total of $3.45K$ GPU hours.

\clearpage

\end{document}